\documentclass{article}
\usepackage{spconf,amsmath,graphicx}
\graphicspath{{/}}
\usepackage{jmei_custom}

\title{Single Index Latent Variable Models for Network Topology Inference}
\name{Jonathan Mei
	and~Jos\'{e}~M.F.~Moura% <-this % stops a space
	\thanks{This work partially funded by NSF grants CCF 1011903 and CCF
		1513936.}}
\address{Carnegie Mellon University\\ Department of Electrical and Computer Engineering\\ 5000 Forbes Avenue\\ Pittsburgh, PA 15213}

\begin{document}
\setlength{\abovedisplayskip}{1pt}
\setlength{\belowdisplayskip}{1pt}
\setlength{\abovedisplayshortskip}{1pt}
\setlength{\belowdisplayshortskip}{1pt}
%\ninept
%
\maketitle
\begin{abstract}
	A semi-parametric, non-linear regression model in the presence of latent variables is applied towards learning network graph structure. These latent variables can correspond to unmodeled phenomena or unmeasured agents in a complex system of interacting entities. This formulation jointly estimates non-linearities in the underlying data generation, the direct interactions between measured entities, and the indirect effects of unmeasured processes on the observed data. The learning is posed as regularized empirical risk minimization. Details of the algorithm for learning the model are outlined. Experiments demonstrate the performance of the learned model on real data.
\end{abstract}
\section{Introduction}

Graphs are a central tool for representing interpretable networks of relationships between interacting entities that generate large amounts of data. Learning network structure while combatting the effects of noise can be achieved via sparse optimization methods, such as regression (Lasso)~\cite{tibshirani_regression_1996} and inverse covariance estimation~\cite{friedman_sparse_2008}. In addition, the extension to time series via vector autoregression~\cite{bolstad_causal_2011,basu_regularized_2015} yields interpretations related to causality~\cite{granger_investigating_1969,mei_signal_2017}. In each of these settings, estimated nonzero values correspond to actual relations, while zeros correspond to absence of relations.

However, we are often unable to collect data to observe all relevant variables, and this leads to observing relationships that may be caused by common links with those unobserved variables. Hidden variables can be fairly general: they can be underlying trends in the data, or the effects of a larger network on an observed subnetwork. For example, one year of daily temperature measurements across a country could be related through a graph based on geographical and meteorological features, but all exhibit the same significant trend due to the changing seasons. We have no single sensor that directly measures this trend. In the literature, a standard pipeline is to de-trend the data as a preprocessing step, and then estimate or use a graph to describe the variations of the data on top of the underlying trends~\cite{sandryhaila_discrete_2013,sandryhaila_discrete_2014,mei_signal_2017}.

Alternatively, attempts have been made to capture the effects of hidden variables via sparse plus low-rank optimization~\cite{chandrasekaran_latent_2012}. This has been extended to time series~\cite{jalali_learning_2011}, and even to a non-linear setting via Generalized Linear Models (GLMs)~\cite{bahadori_fast_2013}. What if the form of the non-linearity (link function) is not known? Regression using a GLM with an unknown link function is also known as a Single Index Model (SIM). Recent results have shown good performance when using SIMs for sparse regression~\cite{ganti_learning_2015}.

Current methods impose a fixed (non-)linearity, assume the absence of any underlying latent variables, perform separate pre-processing or partitioning in an attempt to remove or otherwise explicitly handle such latent variables, or take some combination of these steps.
To address all of these issues, we present a model with a non-linear function applied to a linear argument that captures the effects of latent variables. Thus, we apply the Single Index Latent Variable (SILVar) model~\cite{mei_silvar:_2018}, which uses the SIM in a sparse plus low-rank  optimization setting to enable general, interpretable multi-task regression in the presence of unknown non-linearities and unobserved variables. That is, we examine SILVar as a tool for uncovering hidden relationships buried in data.

First, we introduce the SILVar model in Section~\ref{sec:SILVar}. Then, we outline the numerical procedure for learning the SILVar model in Section~\ref{sec:learn_SILVar}. Finally, we demonstrate the performance via experiments on synthetic and real data in Section~\ref{sec:exp}.

\section{Single Index Latent Variable Models}
\label{sec:SILVar}
In this section, we build the Single Index Latent Variable (SILVar) model from fundamental concepts. We extend the single index model (SIM)~\cite{ichimura_semiparametric_1993} to the multivariate case and account for effects from unmeasured latent variables in the linear parameter. Let $\y_i=\left(y_{1i} \; \ldots \; y_{mi}
\right)^\top$, $\g(\x)=\left(g_1(x_1) \; \ldots \; g_m(x_m) \right)^\top$, $\a_j\in\Rbb^{p}$ for $j=1,\ldots m$, and $\A=\left(\a_1 \; \ldots \; \a_m\right)^\top$. The multivariate SIM model is parameterized by 1) a non-linear link function $g=\nabla G$ where  $G$ is a closed, convex, differentiable, invertible function; and 2) a matrix $\A\in\Rbb^{m\times p}$. Consider the vectorization,
\begin{equation}
\label{eq:lat_vars}
\begin{aligned}
\Ebb \left[y_{ji} | \x_i \right] = g_j \left(\a_j^\top \x_i \right)
\Rightarrow \Ebb \left[\y_i | \x_i \right] = \g \left(\A \x_i \right).
\end{aligned}
\end{equation}
For the remainder of this paper, we make an assumption that all $g_j=g$ for notational simplicity, though the formulations readily extend to the case where $g_j$ are distinct.

We propose the SILVar model,
{\small\begin{equation}
\label{eq:SILVar}
\what{\y} = \what{\g}\left((\what{\A}+\what{\mathbf{L}}) \x \right),
\end{equation}
}where we have explicitly split the linear parameter $\A$ from before into $\A$ and $\mathbf{L}$ such that $\A$ is a sparse matrix, including but not limited to a graph adjacency, and $\mathbf{L}$ is a low-rank matrix ($\textrm{rank}(\mathbf{L})\le r+1$), capturing the indirect effects of a small number $r\ll p$ of unmeasured latent variables on the observed data, as introduced in~\cite{chandrasekaran_latent_2012}. However, in the presence of the non-linearity, it is not obvious that this low-rank representation should faithfully correspond to the latent variables as intended. Luckily, it does~\cite{mei_silvar:_2018}. Letting 
{\small
	\begin{equation}
	\label{eq:SILVar_opt_obj}
	\begin{aligned}
	\what{F}_3(\Y,\X,g,\A) \!\!=\!\! \frac{1}{n}\!\sum\limits_{i=1}^{n}\!\left[ \sum\limits_{j=1}^{m}\!\left[G_{\!*\!}\left(y_{ji} \!\right) \!+\! G\left(\a_j^\top\!\x_i \!\right)\right] \!\!-\! \y_i^\top \!\left(\!\A \x_i\!\right)\! \right],
	\end{aligned}
	\end{equation}
}we learn the model using the optimization problem,
{\small\begin{align}
\label{eq:SILVar_opt}
(\what{g},\what{\A},\what{\mathbf{L}}) = & \argmin[g,\A,\mathbf{L}] \what{F}_3(\Y,\X,g,\A+\mathbf{L}) +h_1(\A)+h_2(\mathbf{L}) \nonumber  \\ & \quad \textrm{s.t. } g=\nabla G \in\Ccal^{1},
\end{align}
}where $h_1$ and $h_2$ are regularizers on $\A$ and $\mathbf{L}$ respectively, and with the set $\Ccal^{1}=\{g : \forall y>x,\; 0 \le g(y)-g(x) \le (y-x) \}$ of monotonic increasing $1$-Lipschitz functions. We impose this functional constraint for uniqueness and conditioning of the solution. A natural choice for $h_2$ would be $h_2(\mathbf{L})=\lambda_2\|\mathbf{L}\|_*$ the nuclear norm since $\mathbf{L}$ is approximately low rank due to the influence of a relatively small number of latent variables. We may choose different forms for $h_1$ depending on our assumptions about the structure of $\A$. For example, if $\A$ is sparse, we may use $h_1(\A)=\lambda_1\|\textrm{v}(\A)\|_1$, the $\ell_1$ norm applied element-wise to the vectorized $\A$ matrix. This is a ``sparse and low-rank'' model, which has been shown under certain geometric incoherence conditions to be identifiable~\cite{chandrasekaran_latent_2012}.

\section{Efficiently Learning SILVar Models}
\label{sec:learn_SILVar}
In this section, we describe the algorithm for learning the SILVar model. Surprisingly, the pseudo-likelihood functional $\what{F}_3$ used for learning the SILVar model in~\eqref{eq:SILVar_opt_obj} is jointly convex in $g$, $\A$, and $\mathbf{L}$ ~\cite{mei_silvar:_2018}. This convexity is enough to ensure that the learning can converge and be computationally efficient.

\subsection{Lipschitz Monotonic Regression}
\label{sec:bg:LMR}
The estimation of $g$ with the objective function including terms $G$ and $G_*$ appears to be an intractable calculus of variations problem. However, there is a marginalization technique that avoids estimating functional gradients with respect to $G$ and $G_*$~\cite{acharyya_parameter_2015}. The technique utilizes Lipschitz monotonic regression (LMR) as a subproblem, for which fast algorithms exist~\cite{mei_silvar:_2018}.

Given ordered pairs $\{x_i,y_i\}$ and additive noise $w_i$, let $x_{[j]}$ denote the $j^{th}$ element of the $\{x_i\}$ sorted in ascending order. Then LMR is described by the problem,
{\small
	\begin{equation}
	\label{eq:Lip_mon_reg}
	\begin{aligned}
	\what{\g} \overset{\Delta}{=} \textrm{LMR}(\y,\x) &= \argmin[\g] \sum\limits_{i=1}^{n}\;\left( g(x_i) - y_i \right)^2\\
	& \textrm{s.t. } 0 \le g\left(x_{[j+1]}\right) \!-\! g\left(x_{[j]}\right) \le x_{[j+1]} \!-\! x_{[j]} \\
	&\qquad\qquad \textrm{for } j=1,\ldots, n-1,
	\end{aligned}
	\end{equation}
}which treats $\{y_i\}$ as noisy observations of a function $g$ indexed by $x$, sampled at points $\{x_i\}$. 

\subsection{Learning SILVar Models}
Algorithm~\ref{alg:SILVar} describes the basic learning procedure for the SILVar model and details the gradient computations while assuming a proximal operator is given.
\begin{algorithm}
	\caption{Single Index Latent Variable (SILVar) Learning}
	\label{alg:SILVar}
	\begin{algorithmic}[1]
		\State Initialize $\what{\A}=\0$, $\what{\mathbf{L}}=\0$
		\While{not converged} Proximal Methods
		\State Computing gradients:
		\vspace{-0.3cm}
		\begin{equation*}
		\begin{aligned}
		\boldsymbol{\Theta} &\leftarrow (\what{\A}+\what{\mathbf{L}})\X\\
		\what{\g} &\leftarrow \textrm{LMR}(\textrm{v}(\Y),\textrm{v}(\boldsymbol{\Theta}))  \quad \textrm{from~\eqref{eq:Lip_mon_reg}}\\
		\nabla_{\A} F_3=\nabla_{\mathbf{L}} F_3&=\sum\limits_{i=1}^n\left( \what\g(\boldsymbol{\theta}_i)-\y_i \right)\x_i^\top
		\end{aligned}
		\end{equation*}
		\vspace{-0.3cm}
		\EndWhile\\
		\Return $(\what{\g},\A,\mathbf{L})$
	\end{algorithmic}
\end{algorithm}

Under certain assumptions~\cite{mei_silvar:_2018}, the solution to the optimization problem~\eqref{eq:SILVar_opt_obj} can be shown to achieve good performance relative to problem parameters including sparsity/rank of linear parameters $\A$ and $\mathbf{L}$ and the magnitude of the effect of latent variables.\vspace{-0.3cm}

\section{Experiments}
\label{sec:exp}
We study the performance of the algorithm via simulations on real data. In these experiments, we show two different regression settings under which the SILVar model can be applied.

\subsection{Temperature Data}
In this setting, we wish to learn the graph capturing relations between the weather patterns at different cities. The data is a real world multivariate time series consisting of daily temperature measurements (in $^\circ$F) for $365$ consecutive days during the year $2011$ taken at each of $150$ different cities across the continental USA.

Previously, the analysis on this dataset has been performed by first fitting with a $4$th order polynomial and then estimating a sparse graph from an autoregressive model using a known link function $g(x)=x$ assuming Gaussian noise~\cite{mei_signal_2017}. 

Here, we fit the time series using a $2$nd order AR SILVar model with regularizers for group sparsity $h_1(\A)=\lambda_1\sum\limits_{i,j}\left\|\left(a^{(1)}_{ij} \ldots a^{(M)}_{ij}\right)\right\|_2$ where $a_{ij}^{(m)}$ is the $ij$ entry of matrix $\A^{(m)}$, and nuclear norm $h_2(\mathbf{L})=\lambda_2\sum\limits_{i=1}^{M}\left\|\mathbf{L}^{(i)}\right\|_*$. 

\begin{figure}[t!]
	\begin{subfigure}[t]{0.925\columnwidth}
		\includegraphics[width=\columnwidth]{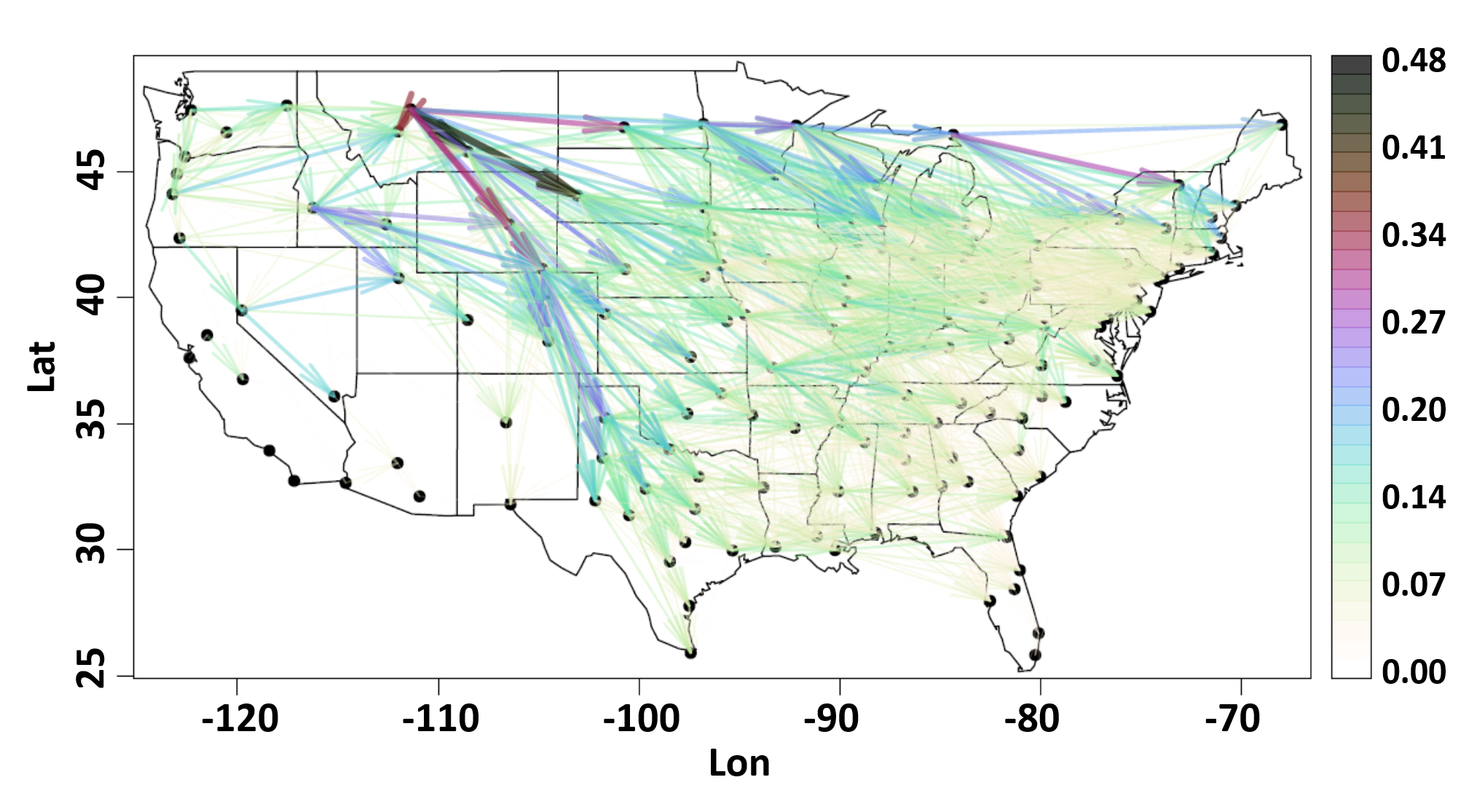}~\vspace*{-.3cm}
		\caption{Weather graph learned using SILVar}
		\label{fig:weather_graphs:silvar}
	\end{subfigure}
	\begin{subfigure}[t]{0.925\columnwidth}
		\includegraphics[width=\columnwidth]{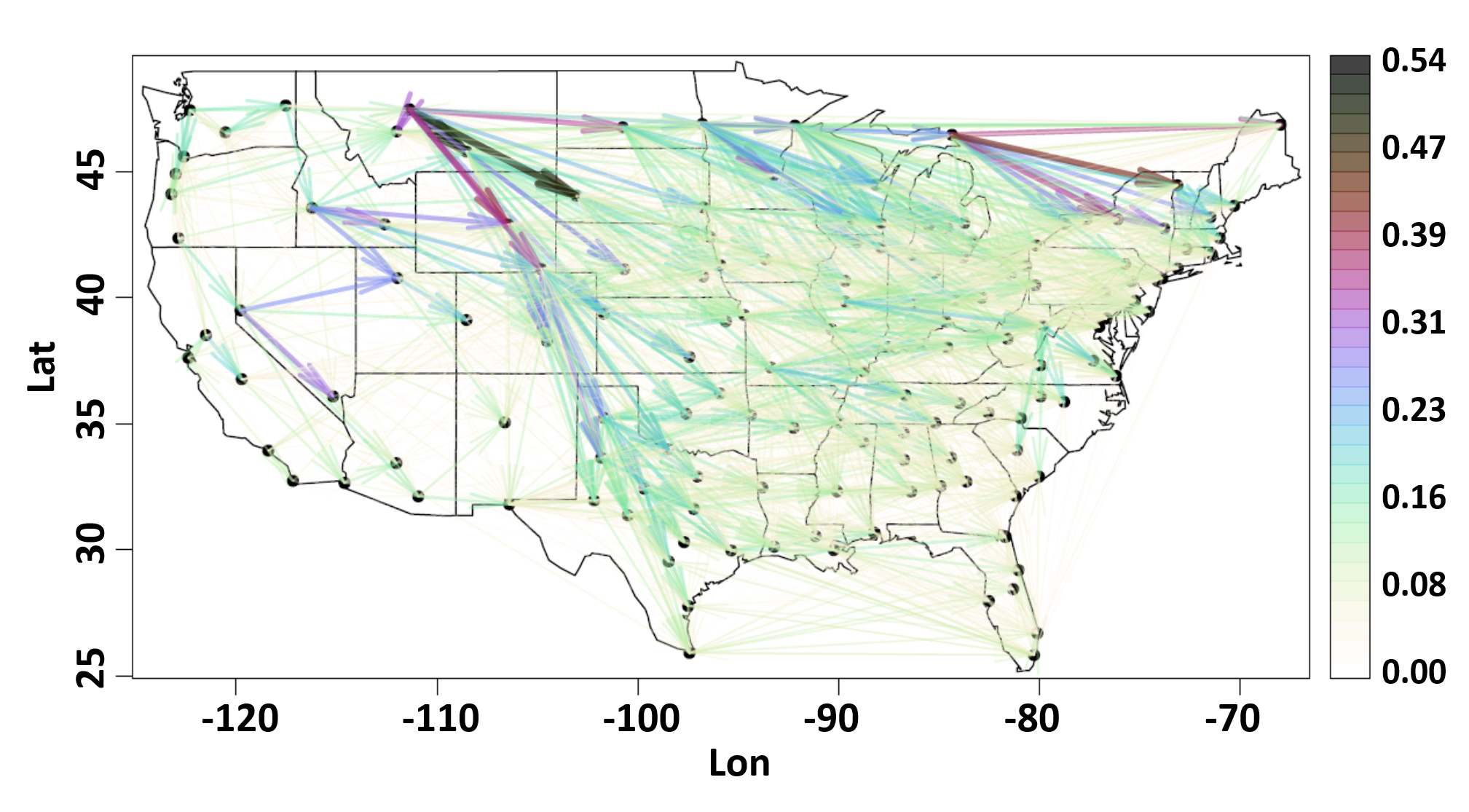}~\vspace*{-.3cm}
		\caption{Weather graph learned using Sp. SIM (without low-rank)}
		\label{fig:weather_graphs:nolat}
	\end{subfigure}~\vspace*{-.3cm}
	\caption{Learned weather stations graphs}
	\label{fig:weather_graphs}		
\end{figure}

Figure~\ref{fig:weather_graphs} compares two networks $\what{\A}'$ estimated using SILVar and using just sparse SIM without accounting for the low-rank trends, both with the same sparsity level of $12\%$ non-zeros for display purposes, and where $\what{a}'_{ij}=\left\|\left(\what{a}^{(1)}_{ij} \ldots \what{a}^{(M)}_{ij}\right)\right\|_2$.
Figure~\ref{fig:weather_graphs:silvar} shows the network $\what{\A}'$ that is estimated using SILVar. The connections imply predictive dependencies between the temperatures in cities connected by the graph. It is intuitively pleasing that the patterns discovered match well previously established results based on first de-trending the data and then separately estimating a network~\cite{mei_signal_2017}. That is, we see the effect of the Rocky Mountain chain around $-110^\circ$ to $-105^\circ$ longitude and the overall west-to-east direction of the weather patterns, matching the prevailing winds.
In contrast to that of SILVar, the graph estimated by the sparse SIM shown in Figure~\ref{fig:weather_graphs:nolat} on the other hand has many additional connections with no basis in actual weather patterns. Two particularly unsatisfying cities are: sunny Los Angeles, California at $(-118, 34)$, with its multiple connections to snowy northern cities including Fargo, North Dakota at $(-97, 47)$; and Caribou, Maine at $(-68, 47)$, with its multiple connections going far westward against prevailing winds including to Helena, Montana at $(-112, 47)$. These do not show in the graph estimated by SILVar and shown in Figure~\ref{fig:weather_graphs:silvar}.\vspace{-0.3cm}

\subsection{Bike Traffic Data}
The bike traffic data was obtained from HealthyRide Pittsburgh~\cite{noauthor_healthy_2016}. The dataset contains the timestamps and station locations of departure and arrival (among other information) for each of 127,559 trips taken between 50 stations within the city from May 31, 2015 to September 30, 2016, a total of 489 days. 

We consider the task of using the total number of rides departing from and arriving in each location at 6:00AM-11:00AM to predict the number of rides departing from each location during the peak period of 11:00AM-2:00PM for each day. This corresponds to $\Y\in\mathbb{N}_0^{50 \times 489}$ and $\X\in\mathbb{N}_0^{100 \times 489}$, where $\mathbb{N}_0$ is the set of non-negative integers, and $\A,\mathbf{L}\in\Rbb^{50\times 100}$.
We estimate the SILVar model~\eqref{eq:SILVar} and compare its performance against a sparse plus low-rank GLM model with an underlying Poisson distribution and fixed link function $g_{\scriptsize{\textrm{GLM}}}(x)=\log(1+e^x)$. We use $n\in\{60,120,240,360\}$ training samples and compute errors on validation and test sets of size $48$ each, and learn the model on a grid of $(\lambda_S,\lambda_L)\in\left\{10^{i/4} \big| i\in \{-8, -7, \ldots, 11, 12\} \right\}^2$. We repeat this 10 times for each setting, using an independent set of training samples each time. We compute testing errors in these cases for the optimal $(\lambda_S,\lambda_L)$ with lowest validation errors for both SILVar and GLM models.

\begin{figure}[t]	
	\begin{subfigure}[t]{0.45\columnwidth}
		\includegraphics[width=\columnwidth]{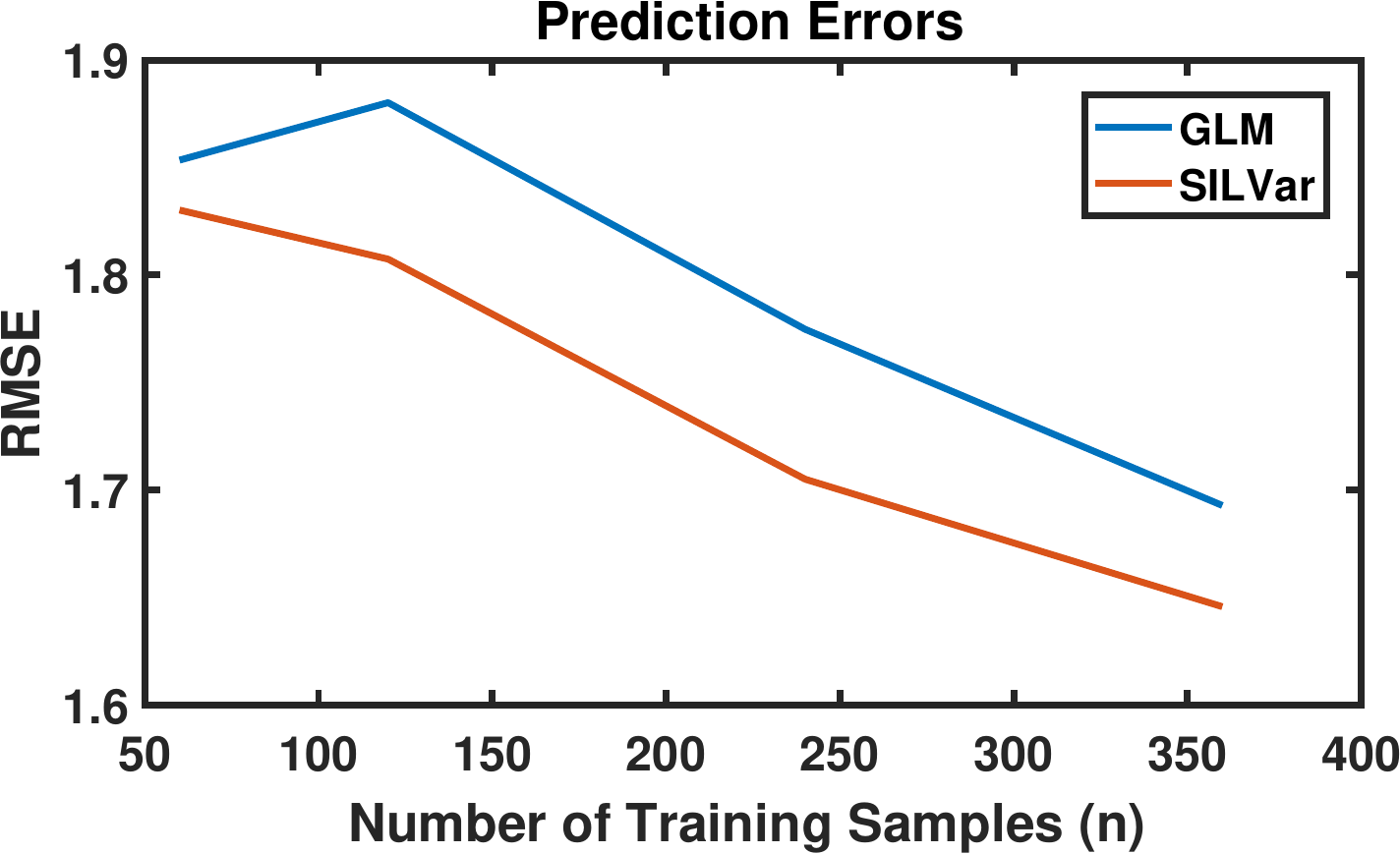}
		\caption{}
		\label{fig:bikes_MSE_link:MSE}
	\end{subfigure}%
	~
	\begin{subfigure}[t]{0.45\columnwidth}
		\includegraphics[width=\columnwidth]{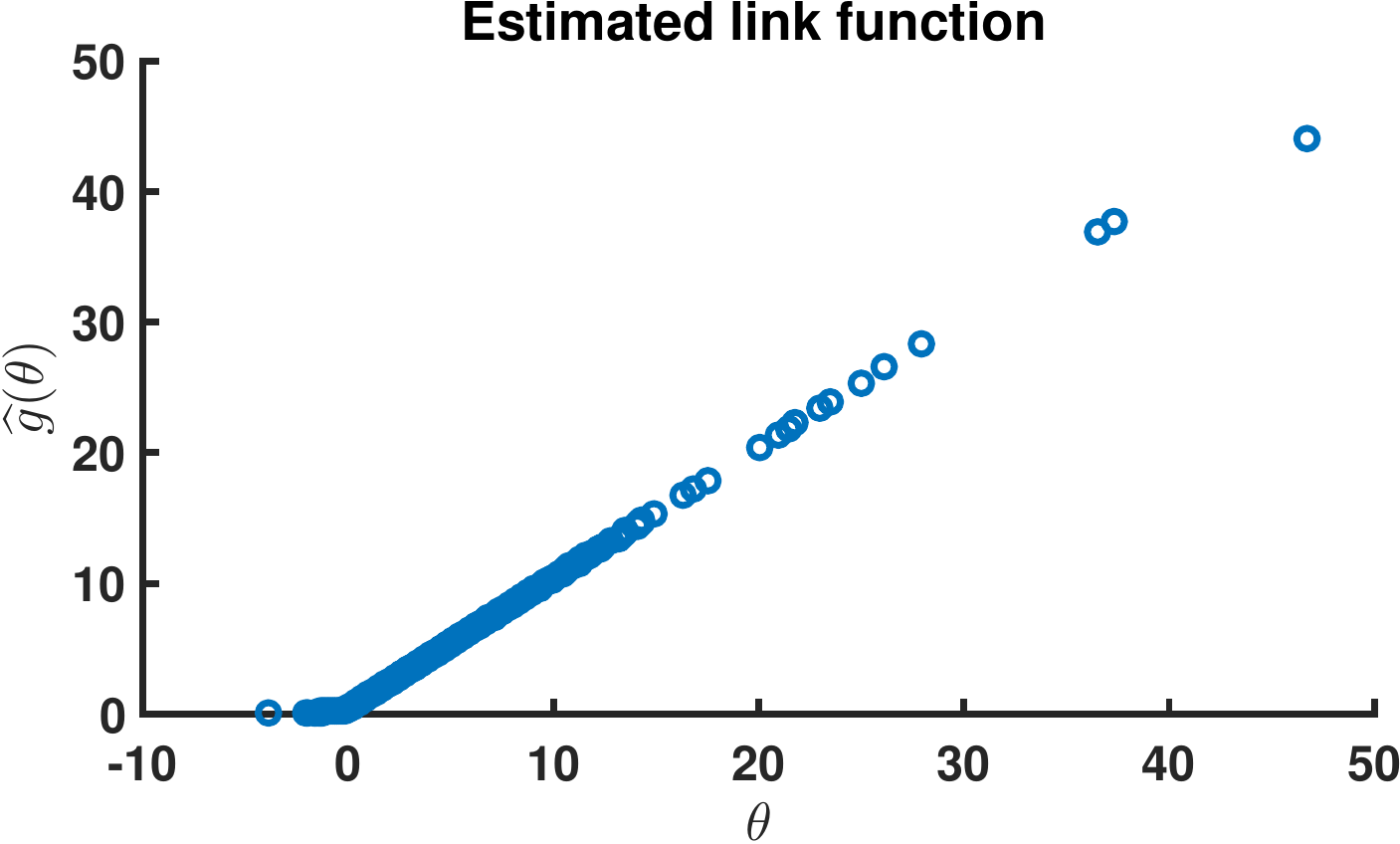}
		\caption{}
		\label{fig:bikes_MSE_link:link}
	\end{subfigure}~\vspace*{-.3cm}
	\caption{(a) Root mean squared errors (RMSEs) from SILVar and Oracle models; (b) Link function learned using SILVar model}
	\label{fig:bikes_MSE_link}
	\vspace*{-.5cm}		
\end{figure}

Figure~\ref{fig:bikes_MSE_link:MSE} shows the test Root Mean Squared Errors (RMSEs) for both SILVar and GLM models for varying training sample sizes, averaged across the 10 trials. We see that the SILVar model outperforms the GLM model by learning the link function in addition to the sparse and low-rank regression matrices. Figure~\ref{fig:bikes_MSE_link:link} shows an example of the link function learned by the SILVar model with $n=360$ training samples, which performs non-negative clipping of the output. This is consistent with the count-valued nature of the data. 

\begin{figure}[t]	
	\includegraphics[width=0.95\columnwidth]{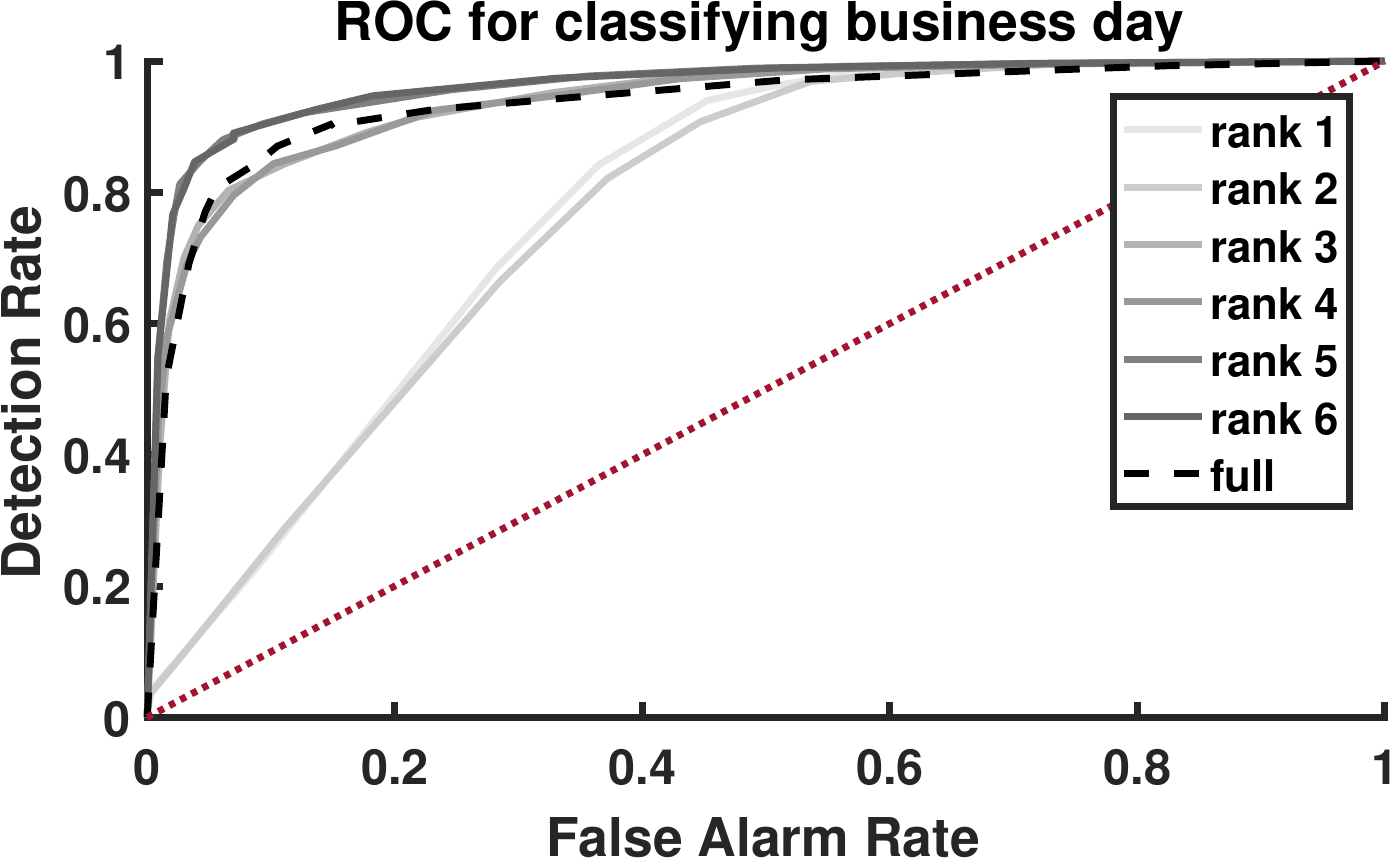}
	\vspace*{-.1cm}
	\caption{Receiver operating characteristics (ROCs) for classifying each day as a business day or non-business day, using the low-rank embedding provided by $\what{\mathbf{L}}$ learned from the SILVar model and using the full data}
	\vspace*{-.5cm}	
	\label{fig:bikes_ROCs}	
\end{figure}

We also demonstrate that the low-rank component of the estimated SILVar model indeed captures unmeasured patterns intrinsic to the data. Naturally, we expect people's behavior and thus traffic to be different on business days and on non-business days. A standard pre-processing step would be to segment the data along this line and learn two different models. However, as we use the full dataset to learn one single model, we hypothesize that the learned $\what{\mathbf{L}}$ captures some aspects of this underlying behavior. To test this hypothesis, we perform the singular value decomposition (SVD) on the optimally learned $\what{\mathbf{L}}=\what{\U}\what{\boldsymbol{\Sigma}}\what{\V}^\top$ for $n=360$ and project the data onto the $r$ top singular components (SC) $\wtil{\X}_r=\what{\boldsymbol{\Sigma}}_r\what{\V}_r^\top\X$. We then use $\wtil{\X}_r$ to train a linear support vector machine (SVM) to classify each day as either a business day or a non-business day, and compare the performance of this lower dimensional feature to that of using the full vector $\X$ to train a linear SVM. If our hypothesis is true then the performance of the classifier trained on $\wtil{\X}_r$ should be competitive with that of the classifier trained on $\X$. We use 50 training samples of $\wtil\X_r$ and of $\X$ and test on the remainder of the data. We repeat this 50 times by drawing a new batch of $50$ samples each time. We then vary the proportion of business to non-business days in the training sample to trace out a receiver operating characteristic (ROC).

In Figure~\ref{fig:bikes_ROCs}, we see the results of training linear SVM on $\wtil{\X}_r$ for $r\in\{1,...,6\}$ and on the full data for classifying business and non-business days. We see that using only the first two SC, the performance is poor. However, by simply taking 3 or 4 SC, the classification performance almost matches that of the full data. Surprisingly, using the top 5 or 6 SC achieves performance greater than that of the full data. This suggests that the projection may even play the role of a de-noising filter in some sense. This classification performance strongly suggests that the low-rank $\what{\mathbf{L}}$ indeed captures the latent behavioral factors in the data.

\begin{figure}[t]	
	\includegraphics[width=0.95\columnwidth]{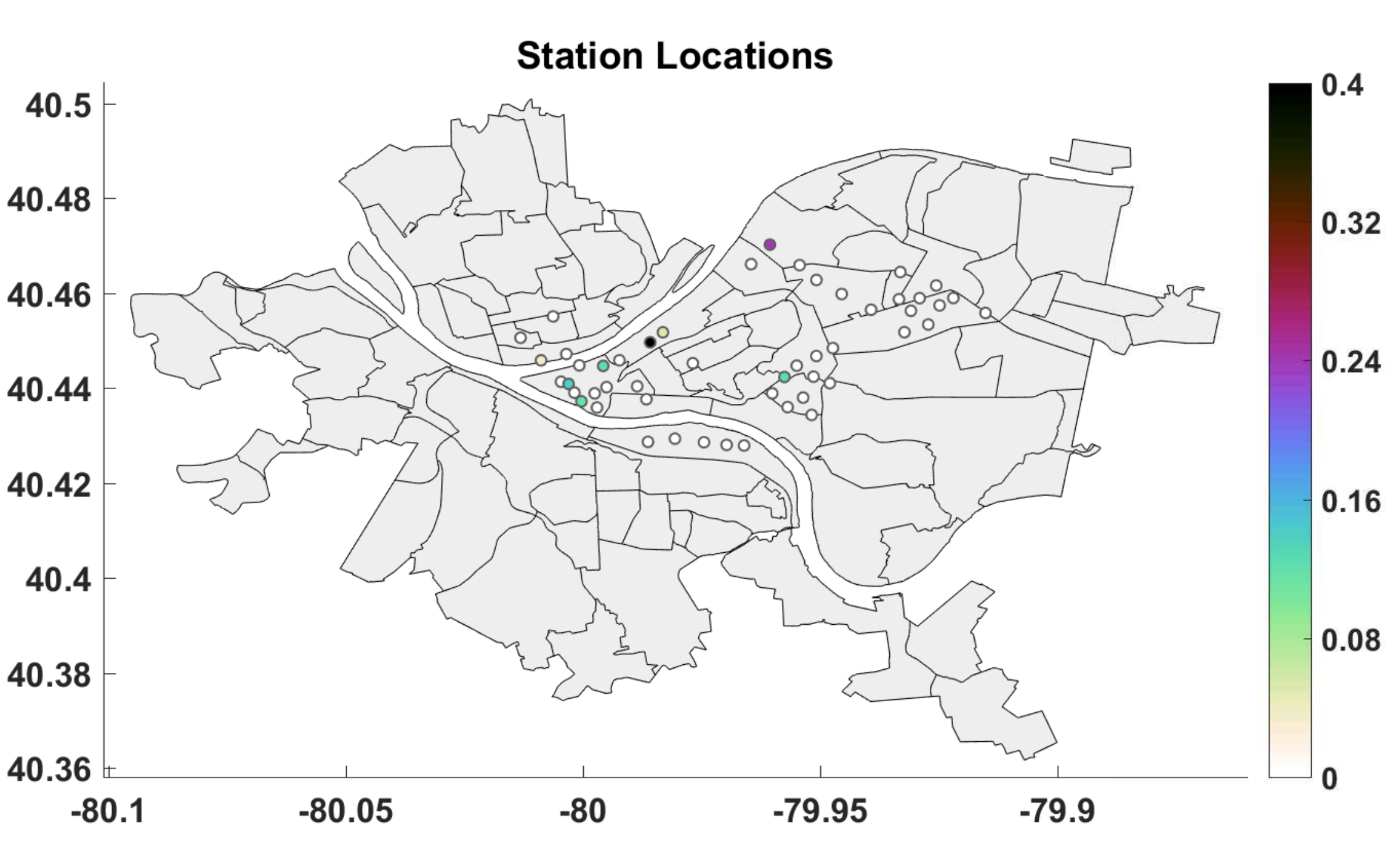}
	\vspace*{-.1cm}
	\caption{Intensities of the self-loop at each station}
	\vspace*{-.5cm}
	\label{fig:bikes_station_location}
\end{figure}
Finally, in Figure~\ref{fig:bikes_station_location}, we plot the diagonal $(i,i)$ entries of the optimal network $\what\A$ at $n=360$, as we find this visualization the most intriguing. This corresponds to locations for which incoming bike rides at 6:00AM-11:00AM are good predictors of outgoing bike rides at 11:00AM-2:00PM, beyond the effect of latent factors such as day of the week. We may expect this to correlate with locations that have restaurants open for lunch service, so that people would be likely to ride in for lunch or ride out after lunch. This is confirmed by observing that these stations are in Downtown (-80,40.44), the Strip District (-79.975, 40.45), Lawrenceville (-79.96, 40.47), and Oakland (-79.96, 40.44), known locations of many restaurants in Pittsburgh. It is especially interesting to note that Oakland, sandwiched between the University of Pittsburgh and Carnegie Mellon University, is included. Even though the target demographic is largely within walking distance, there is a high density of restaurants open for lunch, which may explain its non-zero coefficient. The remainder of the locations with non-zero coefficients $a_{ii}$ are also near high densities of lunch spots, while the other locations with coefficients $a_{ii}$ of zero are largely either near residential areas or near neighborhoods known for dinner or nightlife rather than lunch, such as Shadyside ($x\ge -79.95$) and Southside ($y\le 40.43$)).\vspace{-0.4cm}

\section{Conclusion}
\label{sec:conc}
Data exhibit complex dependencies, and it is often a challenge to deal with non-linearities and unmodeled effects when attempting to uncover meaningful relationships among various interacting entities that generate the data.
We apply the SILVar model to estimating sparse graphs from data under the presence of non-linearities and latent factors or trends. The SILVar model estimates a non-linear link function $g$ as well as structured regression matrices $\A$ and $\mathbf{L}$ in a sparse and low-rank fashion. We outline computationally tractable algorithms for learning the model and demonstrate its performance against existing regression methods on real data sets, namely 2011 US weather sensor network data and 2015-2016 Pittsburgh bike traffic data. We show on the temperature data that the learned $\mathbf{L}$ can account for the effects of underlying trends in time series while $\A$ represents a graph consistent with US weather patterns; and we see that, in the bike data, SILVar outperforms a GLM with a fixed link function, the learned $\mathbf{L}$ encodes latent behavioral aspects of the data, and $\A$ discovers notable locations consistent with the restaurant landscape of Pittsburgh.

\bibliographystyle{IEEEbib}
\bibliography{jmei}

\end{document}